\documentclass[letterpaper, 10 pt, conference]{ieeeconf}
\IEEEoverridecommandlockouts
\usepackage[noadjust]{cite} 
\usepackage{amsmath,amssymb,amsfonts}
\usepackage{algorithm}
\usepackage{algpseudocode}
\usepackage{graphicx}
\usepackage{textcomp}
\usepackage{xcolor}
\usepackage{soul}
\usepackage{xcolor}
\usepackage{subcaption}
\def\BibTeX{{\rm B\kern-.05em{\sc i\kern-.025em b}\kern-.08em
    T\kern-.1667em\lower.7ex\hbox{E}\kern-.125emX}}

\algrenewcommand\algorithmicrequire{\textbf{Input:}}
\algrenewcommand\algorithmicensure{\textbf{Output:}}

\begin{document}
\title{\LARGE \bf Guessing human intentions to avoid dangerous situations in caregiving robots
}

\author{Noé Zapata$^{1}$, Gerardo Pérez$^{1}$, Lucas Bonilla$^{1}$, Pedro Núñez$^{1}$, Pilar Bachiller$^{1}$ and Pablo Bustos$^{1}$
\thanks{$^{1}$All authors are members of RoboLab, University of Extremadura, Spain,  {\tt \small (nzapatac, gperezgonz, lbonillar, pnuntru, pilarb, pbustos)@unex.es}}%
\thanks{This work has been partially funded by the Spanish Ministry of Science and Innovation TED2021-131739-C22, supported by MCIN/AEI/10.13039/501100011033 and the European Union NextGeneration EU/PRTR, by the Spanish Ministry of Science and Innovation PDC2022-133597-C41, by FEDER Project 0124\_EUROAGE\_MAS\_4\_E (2021-2027 POCTEP Program) and by the Spanish Ministry of Science and Innovation PID2022-137344OB-C31 funded by MCIN/AEI/10.13039/501100011033/FEDER, UE.}%
}

\maketitle
\thispagestyle{empty}
\pagestyle{empty}

\begin{abstract}
For robots to interact socially, they must interpret human intentions and anticipate their potential outcomes accurately. This is particularly important for social robots designed for human care, which may face potentially dangerous situations for people, such as unseen obstacles in their way, that should be avoided. This paper explores the Artificial Theory of Mind (ATM) approach to inferring and interpreting human intentions. We propose an algorithm that detects risky situations for humans, selecting a robot action that removes the danger in real time. We use the simulation-based approach to ATM and adopt the "like-me" policy to assign intentions and actions to people. Using this strategy, the robot can detect and act with a high rate of success under time-constrained situations. The algorithm has been implemented as part of an existing robotics cognitive architecture and tested in simulation scenarios. Three experiments have been conducted to test the implementation's robustness, precision and real-time response, including a simulated scenario, a human-in-the-loop hybrid configuration and a real-world scenario. 
\end{abstract}

\section{Introduction}\label{intro}
Caregiving robots can be seen as special social robots that, among other things, react whenever they perceive a dangerous situation for a human. To enable smooth human-robot collaboration, these robots must be aware of their social environment by showing advanced levels of reactivity, interaction, and comprehension \cite{mahdi_survey_2022}. Cultural norms, social cues, human activities, and individual preferences of people around them typically influence social awareness in robots. To be socially adept, robots must perceive their surroundings and occupants, comprehending the situation's social and contextual nuances. 

\begin{figure}[ht]
  \centering
\includegraphics[width=1\linewidth]{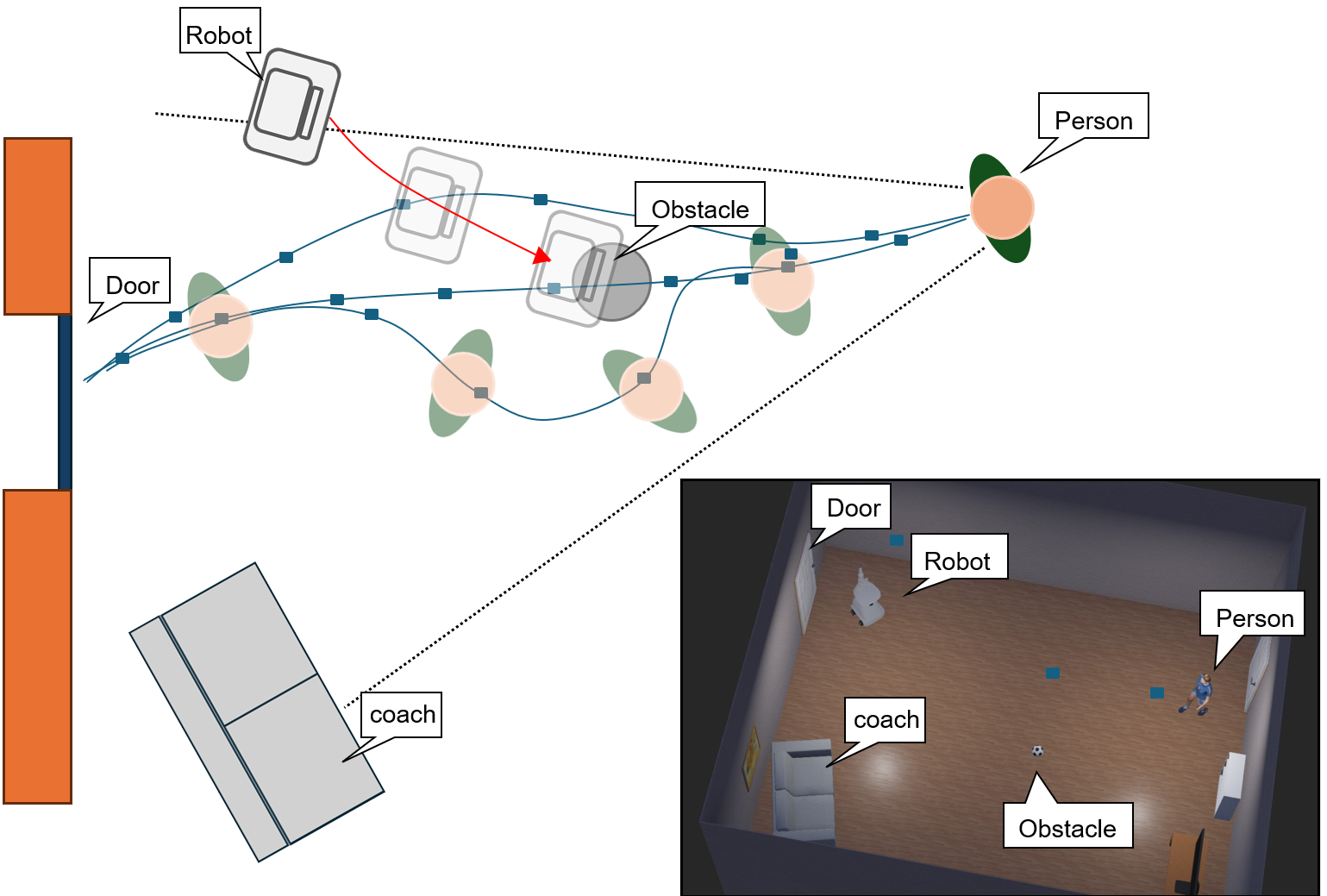}
  \caption{Schematic view of the model. The robot assigns two targets to the human, the door and the couch, and computes one trajectory in which the person collides with the ball. To save the situation, the robot imagines several possible actions to check if one removes imminent danger. The smaller frame shows the scene rendered by the simulator.}
  \label{fig:scenario-1}
\end{figure}

Caregiving robots, in particular, must be able to sense and interpret the ongoing activities of individuals in their environment to anticipate future risk scenarios and adjust their behaviour to mitigate the situation in a socially acceptable manner. To achieve this, robots must be able to predict forthcoming events, foresee potential consequences, and select the most understandable, socially acceptable behaviour to align with human social expectations.
\par

In this paper, we build on previous results in the field of artificial theory of mind (ATM)~\cite{Scassellati2002Theory} and its relation to simulation-based internal models \cite{Winfield-2018}\cite{Blum-2018}. Theory of mind (TM) refers to the ability to predict the actions of oneself and others \cite{Carruthers_Smith_1996}. It provides a plausible explanation of how we can anticipate the behaviour of others in particular circumstances \cite{Winfield-2018}\cite{kennedy_like-me_2009}. Within TM, there are several approaches to explain how the underlying processes work. We follow the simulation theory (ST) variant here, which states that "\textit{... people use imagination, mental pretence, or perspective taking ('putting oneself in the other person's shoes') to determine the mental states of others}" \cite{shanton_simulation_2010}.
Our contribution here is to support further the hypothesis that simulation-based internal models can provide a feasible basis for ATM and, on this basis, to provide a working model and an example of human caregiving in social robots.
\par
We propose a model that assigns intentions to humans using the same machinery used by our robot to attach itself to local goals. This approach is known as a "like-me" simulation~\cite{kennedy_like-me_2009}. 
As part of its existing set of capabilities, our robot \textit{Shadow}\cite{shadow_tool_robot} can select a visually detected object as a target and navigate towards it along an obstacle-free path. 
Accordingly, nearby humans are assigned intentions as links between them and the objects in their environment. The robot interprets these links as the potential to perform an action involving the person and the object, where possible actions are those the robot can already do. 
To understand the consequences of human actions, the robot simulates them using its internal model but adopting a third-person perspective. Following \cite{sanz_d13_nodate}, we assume that the robot understands a phenomenon if it has a model for it and the logical implications in the model correspond to causality in the system. 
To explore this approach further, we have provided the robot with a simple model of itself and its interactions with the environment. 
This model can be run by an internal physics-based simulator and used to explore the simulated consequences of its actions and derive a plan to execute. 
Following this line of reasoning, an interesting way of applying this theory of understanding to the intentions that our model assigns to people is to simulate their possible actions and check whether they have consequences in the environment and whether they can be a security risk. 
If an intention results in a risk to the person, the robot will continue to simulate whether any of its actions would eliminate the risk, and if so, it will carry them out. 
This idea is shown in Figure~\ref{fig:scenario-1}, where a person is assigned two targets, the door and the table, each giving rise to several trajectories. 
As one of them is considered dangerous, the robot imagines an action that could revert the situation. 
The way to confirm that the action will be effective is to simulate the person's trajectory again but with the robot placed at its imagined final destination. As a result, the person acknowledges the presence of the larger volume added by the robot and plans an obstacle-free trajectory.

The paper proceeds by describing other works in ATMs that have been applied to social robots. Our model is then described in detail, followed by the most relevant aspects of its implementation in the CORTEX cognitive architecture \cite{luis-pablo-cortex}\cite{marfil:2020}. To conclude, we will show the results of the simulated and real-world experiments and provide some closing remarks.

\section{Related work}
A pioneering work in Artificial Theory of Mind was the work by Scassellatti \cite{Scassellati2002Theory}, where a detailed framework is proposed and implemented in the humanoid robot Cog, triggering the research in this field. 
An early work that uses simulation to model ATM is \cite{kennedy_like-me_2009}. Their authors conduct the experiments in the embodied version of ACT-R/E. One of the experiments shows that the "like-me" approach can be used to determine where the person is looking when the robot's sensors are not accurate enough. However, they do not attempt to model it as an intention nor to transform it into a plan that can be simulated and evaluated in the model.
Gray et al. \cite{gray_manipulating_2014} describe an ATM experiment in which the robot tries to manipulate the human's beliefs by concatenating two primitives: \textit{action-simulation} and  \textit{mental-state-simulation}. These states correspond with our simulation of human actions before and after the robot intervention. Although the paper presents a sophisticated scheme for recursively modelling beliefs, its goal is to show how embodiment connects agents' mental states rather than applying ATM to a social robot in real-time.

A more recent series of works by A. Winfield and collaborators \cite{Winfield-2018}\cite{Blum-2018} have shown an architecture for a robot with a simulation-based internal model named \textit{consequence engine}. Although their and our approaches are initially similar, their robots live in a 2D grid. Their simulated actions are projections of their state into the future using the existing movement controller. In our model, people are assigned intentions as potential relations with visually detectable objects, which are then transformed into actions. As a result, the robot operates in a more abstract representation akin to adaptation and learning.
Using a different approach, Lemaignan et al.~\cite{lemaignan_artificial_2017} propose an architecture with an ATM module in which the simulation is carried out by a high-level planner using shared plans. In their planning domain, \textit{agents (humans) are first-class entities}. This approach extends ATM into the Theory Theory (TT) variation that differs from using an internal model-based simulator~\cite{mychlmayr_simulation_2002}.
Some other works have explored the relationship between the use of ATM in robots and its effect on people's acceptance or trust \cite{rossi_evaluating_2022}, or its connection with ethics and security~\cite{vanderelst_architecture_2017}.

From a broader perspective, detecting human intentions has gained interest in social robotics and, particularly, in socially aware robot navigation~\cite{singamaneni2023survey}. For instance, in~\cite{FerrerGHS17}, the robot accompanies a person while predicting their intentions to reach a destination point. Human intentions regarding their goal positions are also detected and used in~\cite{KostavelisKGT17}. Their proposal considers places with semantic meaning as potential human goal positions. 
In~\cite{Mavrogiannis2018}, the proposed system detects signals of intentions or preferences over avoidance strategies and makes the agent act towards simplifying people's decision-making. The work in~\cite{Skrzypczyk2021} presents a control scheme for an intelligent wheelchair navigation assistant that detects cooperative and non-cooperative behaviours by comparing the predicted pedestrian positions to the real ones. 
We concur with these works in characterising intentions, but our experiment also explores how intentions are converted into actions and what their consequences may be. Another interesting approach, more in line with the one presented in this work, is proposed in~\cite{cunningham2019}. Given samples from an agent policy, this approach simulates the robot and the agents forward under their assigned policies to obtain predicted states and observation sequences. The robot aims to find an optimal policy based on these observations over a given decision horizon. Their work differs from ours in how the context is represented and how the new goal, i.e. dangerous situations, is detected and integrated into the search for a successful action.

\section{The scenario}
As a test bed for our research, we have created a scenario using the Webots\footnote{https://cyberbotics.com/}{} simulator. As shown in Figure \ref{fig:scenario-1}, a robot observes a person that is about to cross the room, heading towards a couch on the opposite wall. A small object is on the floor, just on the way to the target. The robot has to guess the person's intentions, how they will unfold and, in case of a possible collision, how to act to avoid it. 


The robot is a custom-built unit named Shadow \cite{shadow_tool_robot}, that has a rectangular base with four Mecanum wheels. The robot's main sensors are a  360º camera placed on top of a 3D LiDAR. Both devices are situated in the uppermost part and are co-calibrated. The 3D pointcloud is projected onto the camera image to provide a sparse depth plane that is used to estimate the distance to regions in the image. All objects in the scene are recognisable by the You Only See Once (YOLO)\footnote{We use YoloV8 from Ultralytics. https://www.ultralytics.com/} deep neural network \cite{redmon2016yolo}. There are no other distractors in the room. If the detected element is a person, its skeleton is passed to a second DNN, JointBDOE\footnote{https://github.com/hnuzhy/jointbdoe},  to estimate their orientation \cite{zhou_joint_2023}. All objects detected are assigned a depth coordinate obtained from the LiDAR. Also, they are tracked using the ByteTrack algorithm \cite{zhang2022bytetrack} to assign them a persistent identification tag. 
We define a visual element $e$ as belonging to the set of YOLO recognisable objects $VE$. The stream of detected visual elements is defined as $e_t \in \{V E\}$, where  $t \geq 0 $.
The robot in this scenario can only execute one action, \textit{move\_to($e_i$)}, where $e_i$ is a visually detectable object in the scene. 
It does not have a speaker. 
The action is executed by a local controller consisting of several components: a path planner over a local grid, a model predictive control over the generated trajectory, and a virtual bumper \cite{shadow_tool_robot}. 
No global map is needed to execute visually guided motions. The robot has a known maximum speed and acceleration. 
A new target $e \in \{VE\}$ is sent to the local controller whenever the robot is required to approach a different object in the scene.

\section{The architecture} 
\begin{figure}
    \centering
    \includegraphics[width=0.9\linewidth]{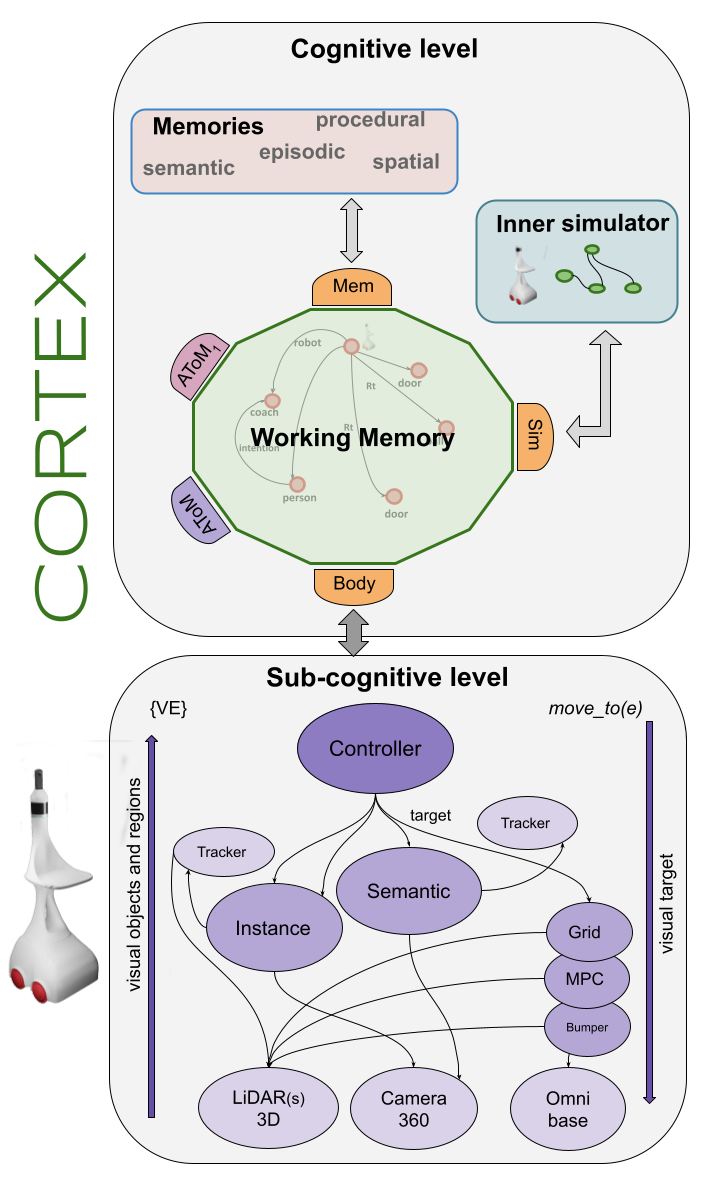}
    \caption{The CORTEX architecture.}
    \label{fig:cortex}
\end{figure}

We use for the experiments a simplified version of the CORTEX robotics cognitive architecture \cite{luis-pablo-cortex}. Figure \ref{fig:cortex} shows a general outline and the elements used in this research.

CORTEX is a multi-agent architecture designed to facilitate the creation of information flows among different types of memories and modules. These flows are pumped by agents. Agents are represented as semicircular forms connected to the working memory $\mathcal{W}$. They can edit it to create and maintain an updated context accessible to the rest of the agents. $\mathcal{W}$ is a distributed graph structure implemented using conflict-free replicated data types (CRDT) \cite{shapiro:inria-00555588}. Its nodes are elements of a predefined ontology, and the edges can encode geometric transformation or logic predicates. Both nodes and edges can store a list of attributes of a predefined type.

The key elements needed to develop this experiment are a) the working memory $\mathcal{W}$; b) the block labelled \textit{sub-cognitive level} that consists of several pipelines of components providing $\mathcal{W}$ with a stream of visually detected elements, and also holds the navigation stack which drives the robot to a visual target; c) the agent connecting the sub-cognitive module with $\mathcal{W}$; d) the internal simulator, based on PyBullet\footnote{https://pybullet.org/} and the agent that interfaces it; and e) the two new agents that implement the ATM and that are described in the next section. More details on the implementation of the working memory can be found in  \cite{garcia_towards_2022}.

\section{The model}
In the "like-me" interpretation of the simulation-based ATM, the robot uses its internal model to attribute intentions to people and to simulate their behaviour, i.e. the robot reads people as robots. 
We define an intention in this scenario as the \textit{engagement that a person establishes with an object in the environment, possibly through its affordances}. 
Intentions are enacted by linking them to one or more of the robot's potential actions and then simulated with its internal model. 
The effects are believed to be causal in the world and can give rise to real actions. 
The context $ctx$ is defined as the content of the working memory $\mathcal{W}$, including all the current robot beliefs about its environment. 
The internal simulator typically runs synchronously with the context, effectively maintaining a copy of the state. 
When the "like-me" process starts, the simulator is temporarily detached, and the people's intentions and the robot's actions can be safely simulated. 
Upon completion, the simulator goes back to synchronisation mode.

We have implemented two algorithms that read and modify the working memory,  $\mathcal{W}$. 
The first agent is activated when people are in $\mathcal{W}$. It will guess and enact their intentions, marking some of them as dangerous if a collision is detected. 
The marked intentions activate the second agent, seeking a feasible robot's intervention that cancels it. 
Both agents communicate through annotations in $\mathcal{W}$. Two predefined lists are provided to simplify the notation: $Gaze$ and $Actions$, which contain the minimum and maximum values for the vertical gaze of a generic human and the list of the robot's actions, respectively.

\begin{algorithm}[h]
\caption{intention guessing and enacting}
\begin{algorithmic}[1]
\Require $W, Gaze, Actions$
\Ensure $W$
\State $ctx \gets getCotext(W)$
\State $People \gets getPeople(ctx)$
\State $Objects \gets getObjects(ctx)$
\ForAll{$p_i \in {People}$}
    \State $Tgts \gets fov(p_i, Objects)$
    \ForAll{$t_j \in Tgts$}
         \ForAll{$act_k \in Actions$}
            \ForAll{$gaze_l \in sample(Gaze)$} 
                \State $int_{ijkl} \gets intention(p_i, t_j, act_k, gaze_l)$
                    \State $c \gets sim(ctx, int_{ijkl})$
                    \State $W \gets addToW(W, (int_{ijkl}, c))$
            \EndFor                
        \EndFor
    \EndFor
\EndFor
\end{algorithmic}
\label{int_guess_alg}
\end{algorithm}

Algorithm \ref{int_guess_alg} starts by accessing the current context $ctx$ provided by the working memory $\mathcal{W}$ and the lists of people and objects of the robot's environment.
For each person, the list of objects entering their field of view ($Tgts$) is obtained using a simple geometric inclusion test over a predefined frustum (line 5).
Each object is assumed to be a potential interaction target for the person. 
Thus, for each person and target, intentions enacted by potential actions must be simulated to predict risky situations.
An intention is considered dangerous if its enactment entails a collision. Since the risk assessment depends on the person's ability to perceive the situation, the person's gaze is also considered during the simulation. We assume the robot cannot estimate it from its position, so the predefined range in $Gaze$ is uniformly sampled (line 8).
According to the person ($p_i$), the object ($t_j$), the action ($act_k$), and the gaze ($gaze_l$), an intention $int_{ijkl}$ is finally generated (line 9).
The simulator is then called to execute that intention given the current context ($ctx$) (line 10). This call implies freezing the current context so changes can be made without interfering with the perception of reality. 
The simulation proceeds by executing action $act_k$ under the constrained access to objects in the scene given by $gaze_l$. 
In this experiment, the robot's only action is \textit{goto(x)}, so the simulator proceeds in two steps: a) a path-planning action to compute a safe route from the person to the target object, in which the occupancy grid is modified to include only the objects in the person's field of view; and b) the displacement of the person along that path in the unmodified scene. 
The first step corresponds to the question "\textit{how does the person move through her environment given the assigned gaze?}" and the second to "\textit{how does the robot imagine from its point of view what the person will do?}''
The path is executed using a copy of the robot's path-following controller. 
The simulation provides the answers to the two questions, returning a flag $c$ signalling the occurrence of a collision. 
Finally, in line 11, the working memory is updated to include the new intention with the attribute $c$, possibly marking a risky situation.

\begin{algorithm}[h]
\caption{Action selection}
\begin{algorithmic}[1]
\Require $W$, $Actions$
\Ensure $W$
\State $ctx \gets getContext(W)$
\State $Intentions \gets getIntentions(ctx)$
\State $Objects \gets getObjects(ctx)$
\ForAll{$int_{ijkl} \in Intentions$ \text{where} $int_{ijkl} \in {Risk}$}
    \ForAll{$act_m \in Actions$}    
        \ForAll{$o_n \in Objects$}  
            \State $robot_{mn} \gets intention(robot, o_n, act_m)$ 
            \State $c \gets sim(ctx, [robot_{mn}, int_{ijkl}])$
            \If {$ \neg c$}
                \State $W \gets addToW(W, robot_{mn})$
            \EndIf
        \EndFor
    \EndFor
\EndFor
\end{algorithmic}
\label{action_sel_alg}
\end{algorithm}

Algorithm \ref{action_sel_alg} runs in a different agent and initiates when an intention is marked as risky in $\mathcal{W}$. The algorithm iterates over the robot's action set, $Actions$, to check if executing any of them removes the risky intention. There is no a priori knowledge of which action to try first or if any of them will succeed. The first loop goes through the intentions marked as dangerous (line 4). 
Then, following a similar procedure as in Algorithm~\ref{int_guess_alg}, the robot assigns to itself intentions to all visible objects. Each robot's intention $robot_{nm}$ along with the person's intention $int_{ijkl}$ are sent for simulation (line 8). 
This call entails the re-enaction of $int_{ijkl}$ but, this time, also considering the robot's action.
If that action makes the collision disappear, it is selected for execution in the real world by adding the corresponding robot's intention to the working memory.

\section{Experiments}
The two proposed algorithms have been implemented as new agents in the architecture and tested with three experiments. The first one consists of a series of simulations on Webots where some free parameters are sampled. The second one introduces uninformed human subjects in the loop to evaluate their reactions to the robot intervention, and the third experiment is performed with a real robot and person.

These experiments are designed to test the ATM model and its implementation as a means of detecting potentially dangerous situations caused by people's intentions. We assume that people always choose the risky path. Thus, we do not try to predict and detect other possible choices. 

\begin{figure}[ht]
    \centering
    \includegraphics[width=0.85\linewidth, height=7.5cm]{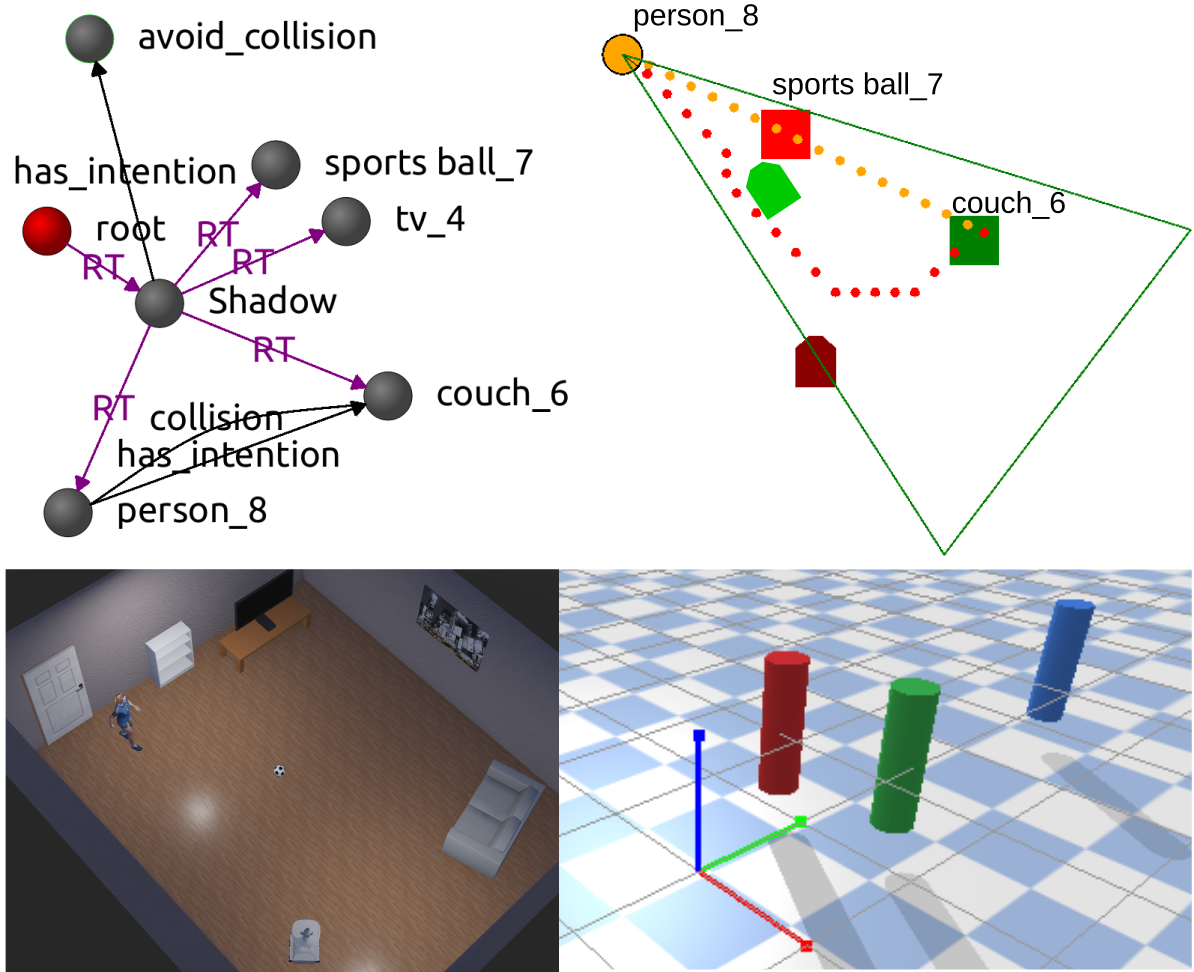}
    \caption{Combined view showing the contents of $\mathcal{W}$ (up-left); a graphical representation of $\mathcal{W}$ with two paths going from the person to the couch (up-right); a zenithal view of the scene as rendered by Webots (down-left); and a 3D view of the internal simulator, PyBullet, with simple geometric forms representing the elements in the scene (down-right)}
    \label{fig:graph-collision}
\end{figure}

The first experiment consists of 180 initialisations of the nominal scenario, letting the robot analyse the situation and decide which action to take. The sampled variables are the positions of the robot, the ball, and the person. The range of positions used in all cases is a $1.5m$ radius. 
The human always follows a direct trajectory towards the couch at a constant speed of $0.5m/s$. Figure \ref{fig:graph-collision} shows the different elements involved in the experiment. The graph in the upper-left shows the current state of $\mathcal{W}$ with the node labelled \textit{Shadow} at the centre. 
The perceived objects and the person are connected to the robot with  \textit{RT} edges that store an $SE(2)$ geometric transformation. 
The rest of the edges are 2-place logic predicates representing perceived relationships between nodes. 
In this example, a \textit{has\_intention} edge connects the person to their intended destination, and a \textit{collision} node signals that they may implement their intention via an unsafe path.
At the top right, the contents of $\mathcal{W}$ are graphically displayed. The person is represented by a yellow circle connected to a green cone, which limits their field of view. Two paths connect the person to the couch. One of the paths crosses the obstacle, represented as a red box, indicating a possible collision. The couch is drawn as a green box. The robot is depicted twice, in red occupying its starting position and in light green occupying the imagined position that will make the person change their path.
The lower left image is a zenithal view of the scene, and the right image shows the PyBullet scene used to simulate the robot and the person traversing the paths.

Table \ref{tab:experimental_results} shows the result of 180 runs with the Webots simulator. 
Of the total number of experiments, $13$ was discarded due to the absence of detected intentions.
Considering the remaining experiments ($167$), the proposed intention-guessing algorithm provides an accuracy rate of $79.64\%$.
The primary source of loss in accuracy is the detection of false positives.
This means that the robot wrongly detects some risky situations that would, in reality, not pose a danger to the person. The primary causes of these false positives are the inaccurate sizes of the shapes representing the person and the obstacles in the internal simulator, and the errors in the estimations of real distances due to noise in the LiDAR measurements.
Nevertheless, the algorithm does not produce false negatives, i.e., all the unsafe intentions are correctly identified by the robot and no risky situation is left unattended.
Both facts can also be derived from the resulting precision and recall values.
Given the consequences of failing to detect jeopardised intentions, recall can be considered more vital than precision.
Taking this into consideration, we measured F1 and F2 scores \cite{sasaki_truth_nodate}.
While the obtained F1 score is moderately high, reaching a value of approximately 0.76, the F2 score indicates a high level of effectiveness.

Regarding time metrics, the mean reaction time, which is the period going from detecting a possible collision to the appearance of an action that can avoid it, is less than $0.75s$. The experiment suggests that a under one second response time is enough for real-time operation in this scenario.
Additionally, the robot can generate an action that prevents the person from colliding in most situations detected as risky.
Although reaction time will grow with the number of objects in the scene, people, and actions with their corresponding free parameters, given some heuristics and pruning strategies discussed in the next section, we believe it can still be kept low enough to work in more complex scenarios. 

\begin{table}[ht]
\centering
\caption{Summary of Experimental Results}
\label{tab:experimental_results}
\begin{tabular}{|l|r|}
\hline
\textbf{Metric} & \textbf{Value} \\
\hline
Total Experiments & 180 \\
Experiments Discarded & 13 \\
Valid Experiments & 167 \\
Accuracy Rate & 79.64\% \\
False Positives & 20.35\% \\
False Negatives & 0\% \\
Precision & 0.61 \\
Recall & 1.0 \\
F-1 Score & 0.76 \\
F-2 Score & 0.89 \\
Reaction time (mean) & 747ms \\
Reaction time (deviation) & 310ms \\
Experiments with Detected Collision and Action Generated & 85.18\% \\
\hline
\end{tabular}
\end{table}

\begin{figure*}[ht!]
    \centering
    \includegraphics[width=0.85\textwidth, height=7cm]{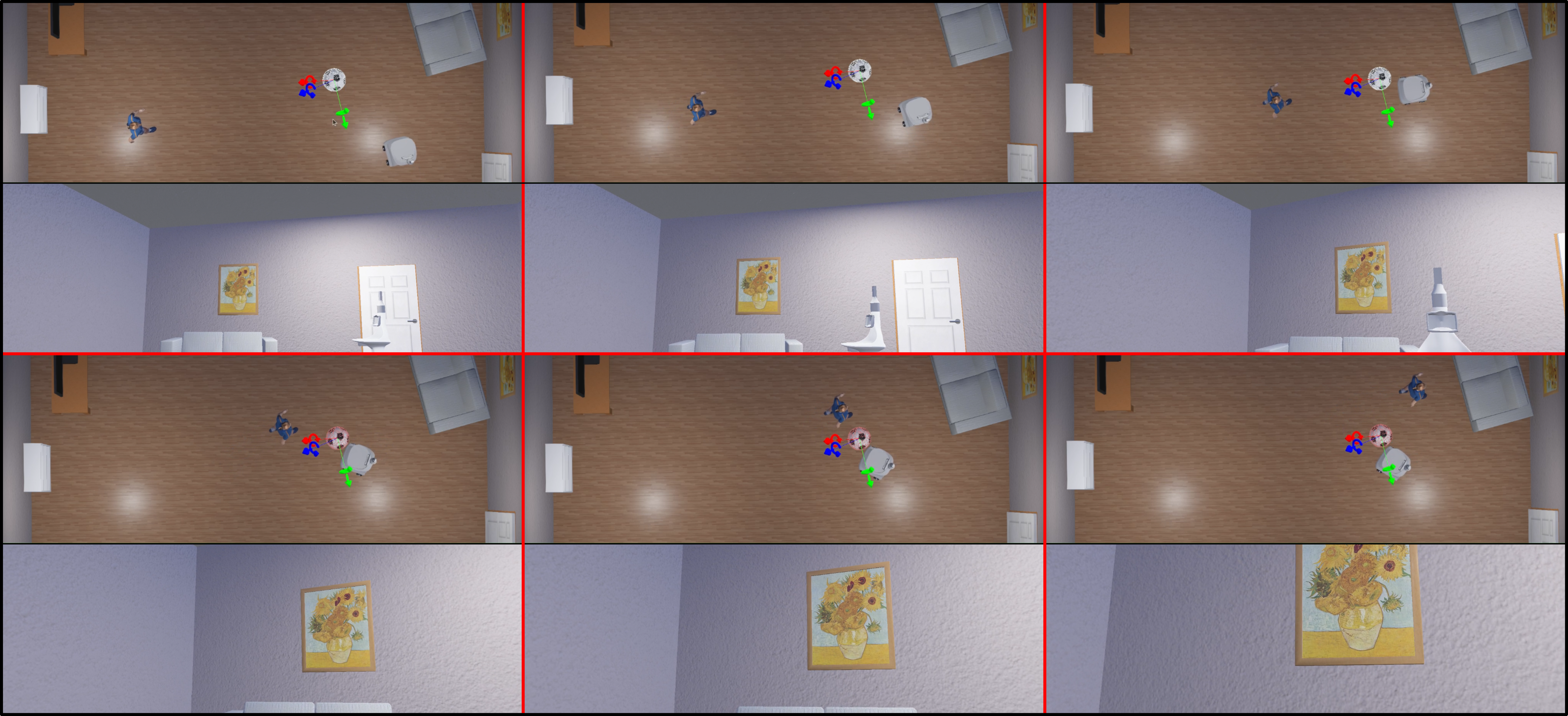}
    \caption{Human-in-the-loop experiment (top left to bottom right). The upper half of each frame is the zenithal view rendered by the Webots simulator. The red and blue axes mark the centre of the scene. The lower half is the view shown to the subject where the ball is missing. When the robot approaches the obstacle, frames 2-3, the subject turns left, overcoming it and safely reaching the couch.}
    \label{fig:human_experiment}
\end{figure*}

A second experiment was conducted to test the reaction of human subjects unfamiliar with the problem to the robot's intervention. This experiment has a human-in-the-loop configuration, where six subjects were instructed to control the person's trajectory in the simulator using a joystick and a view from a virtual camera placed on the avatar. The subjective camera is oriented in a way that prevents the subjects from seeing the ball on the floor. The subject's goal is to reach the couch on the other side of the room, and they are only allowed to play with the joystick for a few seconds. After completing the experiment, all subjects avoided the hidden ball when the robot entered their field of view and was detected. Qualitatively, all subjects generated different trajectories with unequal free margins to the unseen obstacle but completed the task without incidences.
These variations were not considered relevant to the experiment.
Figure \ref{fig:human_experiment} shows a sequence of six frames from one of these trials. 
The series runs from top left to bottom right, and each frame is split into two: the Webots zenithal view of the scene at the top and the subjective view shown to the human subject at the bottom. 
In the sequence, as the robot enters the field of view, the person corrects the trajectory and drives the avatar to the target position.

\begin{figure*}[ht!]
    \centering
    \includegraphics[width=0.85\textwidth, height=7cm]{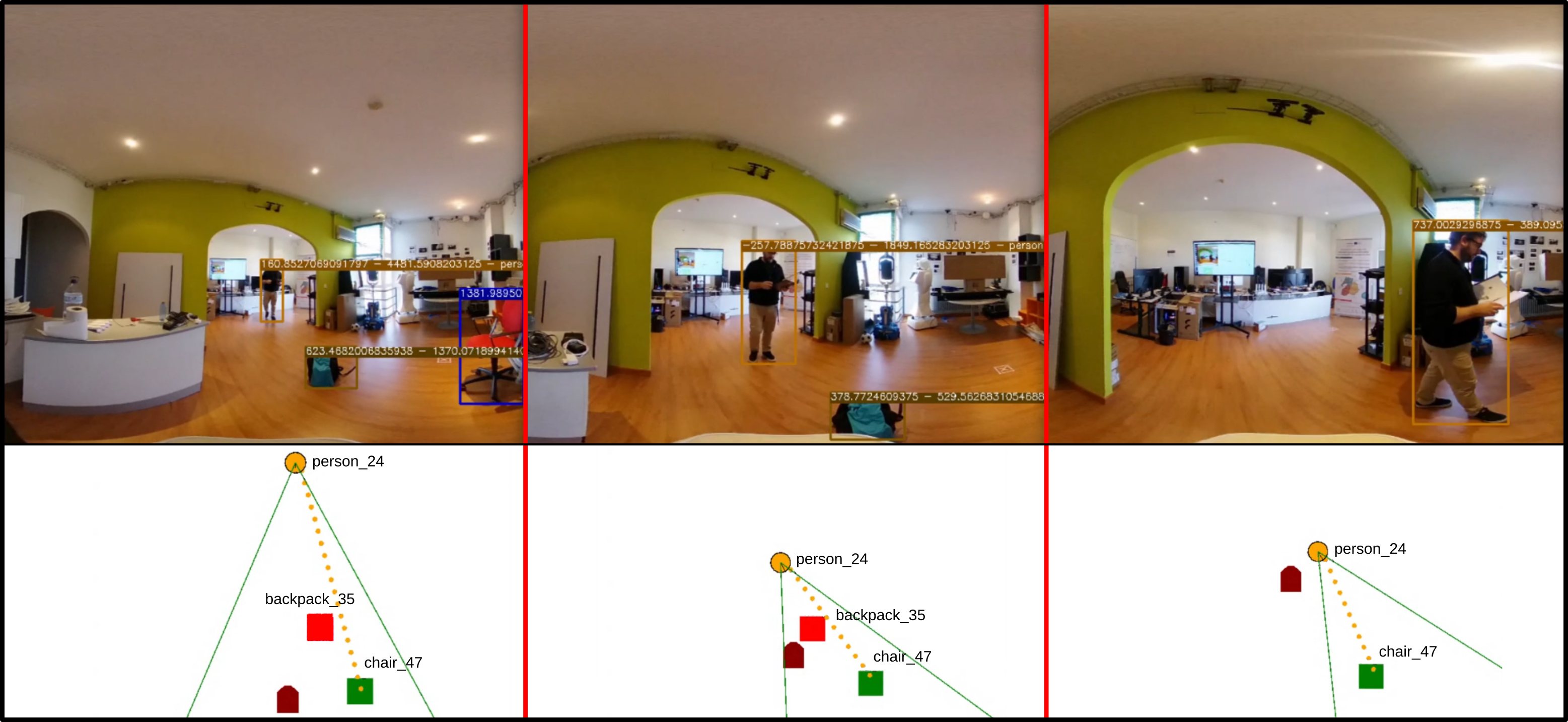}
    \caption{Real world experiment (left to right). The upper half of the frame shows the view from the robot's camera. The lower half shows a schematic view of the working memory with the person represented as a yellow circle, the backpack on the floor as a red square and the target chair as a green square. The robot is coloured dark red. The subject walks distractedly towards the chair (frame 1) and reacts when the robot starts moving (frame 2), changing direction and continuing.}
    \label{fig:real_world_experiment}
\end{figure*}

The third experiment has been performed with the Shadow robot \cite{shadow_tool_robot} in a real scenario. A group of five subjects were instructed to cross the room while reading a paper and heading towards a chair on the other side. A backpack is in the way\footnote{We replaced the football with a backpack because, in the real scenario, YOLO detected it much more confidently.}, and they are instructed to ignore it. They were not briefed about the robot participation. In all five trials, the robot detected the subject and moved close to the unseen obstacle to prevent the person from stumbling. 
The subjects reacted to the robot's intervention with varying degrees of surprise due to the unexpectedness of the event, but all continued on their way to the assigned target. This variety of people's reactions could be used to adapt the internal model according to a comfort metric, resulting in smoother, less startling movements. This idea is commented on in the next section.
Figure \ref{fig:real_world_experiment} shows a three-frame sequence of one of the trials. The lower part of each frame shows a graphic representation of the working memory with the robot in brown heading upwards, the person in yellow heading downwards, the backpack in red, and the chair in green. The line of yellow dots shows the path attributed to the subject during the internal simulation phase.

\section{Discussion and Conclusions}
ATM is slowly gaining momentum as a tool for adding third-view reasoning capabilities to robotics control architecture, but there are still many open issues. 
Our research shows that an internal physics-based simulator embedded in the robot's control architecture could be the key to integrating ATM into the set of cognitive abilities. The current technology allows direct and precise control of the simulator's scene graph and time step. However, this integration relies on the existence of a working memory offering a stable and updated view of the current context based on a hybrid numeric-symbolic representation. This memory relies, in turn, on other modules which provide the necessary connections to the world. Putting all these together implies a considerable development effort and, as we see it, the perspective to further pursue this task is linked to  
three key aspects related to the feasibility of the approach: a) the stability and reliability of the working memory representing the current context in which inferences can be made; b) the combinatorial explosion caused by the nested relationships between participants and the free parameters of their actions; and c) the robustness required for reliable real-time execution in more open scenarios.

The multi-agent CORTEX architecture addresses the first issue to a reasonable amount, with the working memory $\mathcal{W}$ being updated by agents that connect to the subcognitive modules. Although there is ample room for improvement, the results show that the representation is stable and can handle a real-world, real-time situation under fairly realistic conditions. In addition, the underlying data structure, which holds symbolic and numeric values, can be monitored to provide a clear view of the internal reasoning process, facilitating the path to accountability and ethics.  

A second problem is the exponential growth caused by the nested searches that unfold all possible relationships among the robot, its actions, the objects in the scene and the people. This search can be curtailed if we consider ways to exit the loops as soon as possible or even to skip them altogether. Studies in humans suggest that no more than two or three alternative strategies are evaluated, at least without interrupting the ongoing behaviour and resorting to more extended reflection \cite{donoso_human_2014}\cite{kahneman2011thinking}.
A first prune can be achieved by forcing time and space constraints. 
In our experiments, the person-to-collision time limits the admissibility of a robot's action, i.e. displacements that will not arrive in time are discarded. Another limitation comes from the maximum number of people that can be represented in $\mathcal{W}$ simultaneously, which is an architecture parameter. This number could be further reduced by distance to the robot, relative position or some person's attributes that could be relevant in the current context. A further reduction can be achieved if the search ends as soon as an action is found that removes the danger, but more efficiently if the sets are ordered according to some metric or drawn from an a priori distribution learnt from experience. For example, intentions could be assigned to objects according to prior knowledge of the person's habits, and actions could be assigned to intentions biased by the objects' affordances.

The third aspect, robustness, is particularly relevant for robot operation in human-populated scenarios. Robustness depends on the stability of the representation, on the ability to learn from experience the optimal values of actions and objects free parameters, i.e. robot speed, human habits, specific place to stop, time to start, etc., in each context, and on the ability to recover from failures and unforeseen events during the execution. This suggests that to improve the robustness, the control architecture has to be more adaptable at each level of its organisation, integrating learning algorithms that capture the effect of the interventions on humans in specific contexts to fine-tune the free parameters.

An interesting issue arises when more than one person is on the scene, and they have assigned risk intentions. In this case, the robot would face the dilemma of saving one and ignoring the other based. This kind of decision would involve a more complex evaluation of the situation, using a system of values and thus entering the emerging field of social robot ethics \cite{van_maris_new_2021}.

The work presented in this paper is a new step towards an ATM that can be effectively run inside a robotics cognitive architecture. We rely on a working memory that holds a representation stable enough for an inner simulator to take third-view perspectives. In future work, we expect to increase the number and complexity of the robot's actions while keeping the number of combinations small, using algorithms that learn an optimal selection order from experience, anytime design techniques or sorting heuristics from external knowledge sources. 
    
\bibliographystyle{IEEEtran}
\bibliography{bibliography}

\begin{thebibliography}{10}
\providecommand{\url}[1]{#1}
\csname url@samestyle\endcsname
\providecommand{\newblock}{\relax}
\providecommand{\bibinfo}[2]{#2}
\providecommand{\BIBentrySTDinterwordspacing}{\spaceskip=0pt\relax}
\providecommand{\BIBentryALTinterwordstretchfactor}{4}
\providecommand{\BIBentryALTinterwordspacing}{\spaceskip=\fontdimen2\font plus
\BIBentryALTinterwordstretchfactor\fontdimen3\font minus \fontdimen4\font\relax}
\providecommand{\BIBforeignlanguage}[2]{{%
\expandafter\ifx\csname l@#1\endcsname\relax
\typeout{** WARNING: IEEEtran.bst: No hyphenation pattern has been}%
\typeout{** loaded for the language `#1'. Using the pattern for}%
\typeout{** the default language instead.}%
\else
\language=\csname l@#1\endcsname
\fi
#2}}
\providecommand{\BIBdecl}{\relax}
\BIBdecl

\bibitem{mahdi_survey_2022}
\BIBentryALTinterwordspacing
H.~Mahdi, S.~A. Akgun, S.~Saleh, and K.~Dautenhahn, ``A survey on the design and evolution of social robots — {Past}, present and future,'' \emph{Robotics and Autonomous Systems}, vol. 156, p. 104193, Oct. 2022. [Online]. Available: \url{https://www.sciencedirect.com/science/article/pii/S0921889022001117}
\BIBentrySTDinterwordspacing

\bibitem{Scassellati2002Theory}
B.~Scassellati, ``Theory of mind for a humanoid robot,'' \emph{Autonomous Robots}, vol.~12, pp. 13--24, 2002.

\bibitem{Winfield-2018}
\BIBentryALTinterwordspacing
A.~F.~T. Winfield, ``Experiments in artificial theory of mind: From safety to story-telling,'' \emph{Frontiers in Robotics and AI}, vol.~5, 2018. [Online]. Available: \url{https://www.frontiersin.org/articles/10.3389/frobt.2018.00075}
\BIBentrySTDinterwordspacing

\bibitem{Blum-2018}
\BIBentryALTinterwordspacing
C.~Blum, A.~F.~T. Winfield, and V.~V. Hafner, ``Simulation-based internal models for safer robots,'' \emph{Frontiers in Robotics and AI}, vol.~4, 2018. [Online]. Available: \url{https://www.frontiersin.org/articles/10.3389/frobt.2017.00074}
\BIBentrySTDinterwordspacing

\bibitem{Carruthers_Smith_1996}
\emph{Theories of Theories of Mind}.\hskip 1em plus 0.5em minus 0.4em\relax Cambridge University Press, 1996.

\bibitem{kennedy_like-me_2009}
W.~Kennedy, M.~Bugajska, A.~Harrison, and J.~Trafton, ``“{Like}-{Me}” {Simulation} as an {Effective} and {Cognitively} {Plausible} {Basis} for {Social} {Robotics},'' \emph{I. J. Social Robotics}, vol.~1, pp. 181--194, Apr. 2009.

\bibitem{shanton_simulation_2010}
\BIBentryALTinterwordspacing
K.~Shanton and A.~Goldman, ``\BIBforeignlanguage{en}{Simulation theory},'' \emph{\BIBforeignlanguage{en}{WIREs Cognitive Science}}, vol.~1, no.~4, pp. 527--538, Jul. 2010. [Online]. Available: \url{https://wires.onlinelibrary.wiley.com/doi/10.1002/wcs.33}
\BIBentrySTDinterwordspacing

\bibitem{shadow_tool_robot}
A.~Torrejon, N.~Zapata, P.~Núnez, L.~Bonilla, and P.~Bustos, ``Shadow, an accompanying tool robot,'' in \emph{Proceedings of the Workshop on Autonomous Systems 2023}, Aranjuez, Madrid, 2023.

\bibitem{sanz_d13_nodate}
R.~Sanz, M.~Rodriguez, and E.~Aguado, ``\BIBforeignlanguage{en}{D1.3 {Theory} of {Understanding}}.''

\bibitem{luis-pablo-cortex}
\BIBentryALTinterwordspacing
P.~Bustos~García, L.~Manso~Argüelles, A.~Bandera, J.~Bandera, I.~García-Varea, and J.~Martínez-Gómez, ``The cortex cognitive robotics architecture: Use cases,'' \emph{Cognitive Systems Research}, vol.~55, pp. 107 -- 123, 2019. [Online]. Available: \url{http://www.sciencedirect.com/science/article/pii/S1389041717300347}
\BIBentrySTDinterwordspacing

\bibitem{marfil:2020}
R.~Marfil, A.~Romero-Garc{\'e}s, J.~P. Bandera, L.~J. Manso, L.~V. Calderita, P.~Bustos, A.~Bandera, J.~Garc{\'i}a-Polo, F.~Fern{\'a}ndez, and D.~Voilmy, ``Perceptions or actions? grounding how agents interact within a software architecture for cognitive robotics,'' \emph{Cognitive Computation}, vol.~12, pp. 479 -- 497, 2020.

\bibitem{gray_manipulating_2014}
\BIBentryALTinterwordspacing
J.~Gray and C.~Breazeal, ``\BIBforeignlanguage{en}{Manipulating {Mental} {States} {Through} {Physical} {Action}},'' \emph{\BIBforeignlanguage{en}{International Journal of Social Robotics}}, vol.~6, no.~3, pp. 315--327, Aug. 2014. [Online]. Available: \url{https://doi.org/10.1007/s12369-014-0234-2}
\BIBentrySTDinterwordspacing

\bibitem{lemaignan_artificial_2017}
\BIBentryALTinterwordspacing
S.~Lemaignan, M.~Warnier, E.~A. Sisbot, A.~Clodic, and R.~Alami, ``Artificial {Cognition} for {Social} {Human}-{Robot} {Interaction}: {An} {Implementation},'' \emph{Artificial Intelligence}, vol. 247, pp. 45--69, Jun. 2017, publisher: Elsevier. [Online]. Available: \url{https://hal.science/hal-01857498}
\BIBentrySTDinterwordspacing

\bibitem{mychlmayr_simulation_2002}
\BIBentryALTinterwordspacing
``Simulation {Theory} {Versus} {Theory} {Theory}: {Theories} {Concerning} the {Ability} to {Read} {Minds},'' Ph.D. dissertation, Master’s thesis, Leopold-Franzens- Universität Innsbruck., 2002. [Online]. Available: \url{https://www.cyrius.com/publications/michlmayr-tom.pdf}
\BIBentrySTDinterwordspacing

\bibitem{rossi_evaluating_2022}
\BIBentryALTinterwordspacing
A.~Rossi, A.~Andriella, S.~Rossi, C.~Torras, and G.~Alenyà, ``Evaluating the {Effect} of {Theory} of {Mind} on {People}’s {Trust} in a {Faulty} {Robot},'' in \emph{2022 31st {IEEE} {International} {Conference} on {Robot} and {Human} {Interactive} {Communication} ({RO}-{MAN})}, Aug. 2022, pp. 477--482, iSSN: 1944-9437. [Online]. Available: \url{https://ieeexplore.ieee.org/document/9900695}
\BIBentrySTDinterwordspacing

\bibitem{vanderelst_architecture_2017}
D.~Vanderelst and A.~Winfield, ``An architecture for ethical robots inspired by the simulation theory of cognition,'' \emph{Cognitive Systems Research}, vol.~48, May 2017.

\bibitem{singamaneni2023survey}
P.~T. Singamaneni, P.~Bachiller-Burgos, L.~J. Manso, A.~Garrell, A.~Sanfeliu, A.~Spalanzani, and R.~Alami, ``A survey on socially aware robot navigation: Taxonomy and future challenges,'' 2023.

\bibitem{FerrerGHS17}
\BIBentryALTinterwordspacing
G.~Ferrer, A.~Garrell, F.~Herrero, and A.~Sanfeliu, ``Robot social-aware navigation framework to accompany people walking side-by-side,'' \emph{Auton. Robots}, vol.~41, no.~4, pp. 775--793, 2017. [Online]. Available: \url{https://doi.org/10.1007/s10514-016-9584-y}
\BIBentrySTDinterwordspacing

\bibitem{KostavelisKGT17}
I.~Kostavelis, A.~Kargakos, D.~Giakoumis, and D.~Tzovaras, ``Robot's workspace enhancement with dynamic human presence for socially-aware navigation.'' in \emph{ICVS}, ser. Lecture Notes in Computer Science, M.~Liu, H.~Chen, and M.~Vincze, Eds., vol. 10528.\hskip 1em plus 0.5em minus 0.4em\relax Springer, 2017, pp. 279--288.

\bibitem{Mavrogiannis2018}
\BIBentryALTinterwordspacing
C.~I. Mavrogiannis, W.~B. Thomason, and R.~A. Knepper, ``Social momentum: A framework for legible navigation in dynamic multi-agent environments,'' in \emph{Proceedings of the 2018 ACM/IEEE International Conference on Human-Robot Interaction}, ser. HRI '18.\hskip 1em plus 0.5em minus 0.4em\relax New York, NY, USA: Association for Computing Machinery, 2018, p. 361–369. [Online]. Available: \url{https://doi.org/10.1145/3171221.3171255}
\BIBentrySTDinterwordspacing

\bibitem{Skrzypczyk2021}
K.~Skrzypczyk, ``Game against nature based control of an intelligent wheelchair with adaptation to pedestrians' behaviour,'' in \emph{2021 25th International Conference on Methods and Models in Automation and Robotics (MMAR)}, 2021, pp. 285--290.

\bibitem{cunningham2019}
A.~Cunningham, E.~Galceran, D.~Mehta, G.~Ferrer, R.~Eustice, and E.~Olson, \emph{{MPDM}: Multi-policy Decision-Making from Autonomous Driving to Social Robot Navigation}, January 2019, pp. 201--223.

\bibitem{redmon2016yolo}
J.~Redmon, S.~Divvala, R.~Girshick, and A.~Farhadi, ``You only look once: Unified, real-time object detection,'' in \emph{Proceedings of the IEEE conference on computer vision and pattern recognition}, 2016, pp. 779--788.

\bibitem{zhou_joint_2023}
\BIBentryALTinterwordspacing
H.~Zhou, F.~Jiang, J.~Si, and H.~Lu, ``Joint {Multi}-{Person} {Body} {Detection} and {Orientation} {Estimation} via {One} {Unified} {Embedding},'' Mar. 2023, arXiv:2210.15586 [cs] version: 2. [Online]. Available: \url{http://arxiv.org/abs/2210.15586}
\BIBentrySTDinterwordspacing

\bibitem{zhang2022bytetrack}
Y.~Zhang, P.~Sun, Y.~Jiang, D.~Yu, F.~Weng, Z.~Yuan, P.~Luo, W.~Liu, and X.~Wang, ``Bytetrack: Multi-object tracking by associating every detection box,'' in \emph{European Conference on Computer Vision}.\hskip 1em plus 0.5em minus 0.4em\relax Springer, 2022, pp. 1--21.

\bibitem{shapiro:inria-00555588}
\BIBentryALTinterwordspacing
M.~Shapiro, N.~Pregui{\c c}a, C.~Baquero, and M.~Zawirski, ``{A comprehensive study of Convergent and Commutative Replicated Data Types},'' {Inria -- Centre Paris-Rocquencourt ; INRIA}, Research Report RR-7506, Jan. 2011. [Online]. Available: \url{https://inria.hal.science/inria-00555588}
\BIBentrySTDinterwordspacing

\bibitem{garcia_towards_2022}
\BIBentryALTinterwordspacing
J.~C. García, P.~Bachiller, P.~Bustos, and P.~Núñez, ``Towards the design of efficient and versatile cognitive robotic architecture based on distributed, low-latency working memory,'' in \emph{2022 {IEEE} {International} {Conference} on {Autonomous} {Robot} {Systems} and {Competitions} ({ICARSC})}, Apr. 2022, pp. 9--14. [Online]. Available: \url{https://ieeexplore.ieee.org/document/9784798}
\BIBentrySTDinterwordspacing

\bibitem{sasaki_truth_nodate}
Y.~Sasaki, ``\BIBforeignlanguage{en}{The truth of the {F}-measure}.''

\bibitem{donoso_human_2014}
M.~Donoso, A.~Collins, and E.~Koechlin, ``Human cognition. {Foundations} of human reasoning in the prefrontal cortex,'' \emph{Science (New York, N.Y.)}, vol. 344, May 2014.

\bibitem{kahneman2011thinking}
D.~Kahneman, \emph{Thinking, fast and slow}.\hskip 1em plus 0.5em minus 0.4em\relax New York: Farrar, Straus and Giroux, 2011.

\bibitem{van_maris_new_2021}
A.~van Maris, N.~Zook, S.~Dogramadzi, M.~Studley, A.~Winfield, and P.~Caleb-Solly, ``\BIBforeignlanguage{en}{A {New} {Perspective} on {Robot} {Ethics} through {Investigating} {Human}–{Robot} {Interactions} with {Older} {Adults}},'' \emph{\BIBforeignlanguage{en}{Applied Sciences}}, vol.~11, no.~21, p. 10136, Jan. 2021, number: 21.

\end{thebibliography}

\end{document}